\documentclass[sigconf]{acmart}

\AtBeginDocument{%
	}

\setcopyright{rightsretained}
\copyrightyear{2024}
\acmDOI{}
\acmISBN{}

\acmConference[MRR 2024]{the 2024 SIGIR Workshop on Multimodal Representation and Retrieval}{July 18, 2024}{Washington, DC}

\begin{document}
	
	\title{Video Enriched Retrieval Augmented Generation Using Aligned Video Captions}
	
	\author{Kevin Dela Rosa}
	\affiliation{%
		\institution{Snap Inc.}
		\city{Santa Monica}
		\state{California}
		\country{USA}
	}
	\email{kevd1337@gmail.com}

	\begin{abstract}
		In this work, we propose the use of "aligned visual captions" as a mechanism for integrating information contained within videos into retrieval augmented generation (RAG) based chat assistant systems. These captions are able to describe the visual and audio content of videos in a large corpus while having the advantage of being in a textual format that is both easy to reason about \& incorporate into large language model (LLM) prompts, but also typically require less multimedia content to be inserted into the multimodal LLM context window, where typical configurations can aggressively fill up the context window by sampling video frames from the source video. Furthermore, visual captions can be adapted to specific use cases by prompting the original foundational model / captioner for particular visual details or fine tuning. In hopes of helping advancing progress in this area, we curate a dataset and describe automatic evaluation procedures on common RAG tasks.
	\end{abstract}

		
		
		\begin{teaserfigure}
			\includegraphics[width=\textwidth]{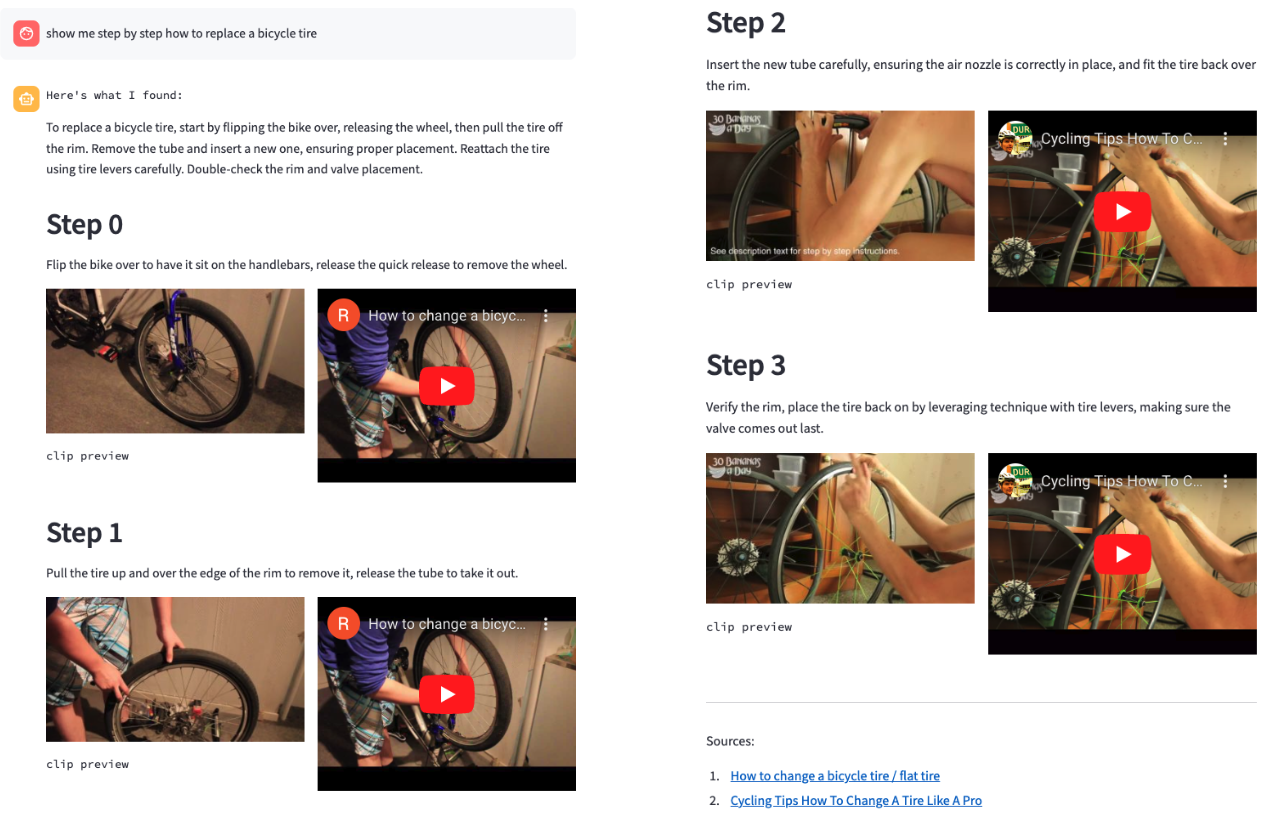}
			\caption{Sample output of chat assistant app leveraging retrieved video content stored in the form of aligned video captions} 
			\label{fig:teaser}
		\end{teaserfigure}
		
		
		\maketitle
		
		\section{Introduction}
		\label{sec:intro}
		
		Video content in the forms of YouTube shorts, TikToks, Instagram Reels or the like are quickly becoming many people's main form of content ingestion online. At the same time, following the initial deluge of work on large language models there has been a recent surge in work to understand videos, with big corporate systems like OpenAI GPT-4 Vision and Google Gemini incorporating basic image and video chatting capabilities into their chat AI applications, as well as academic systems like \cite{Maaz2023VideoChatGPT}  \cite{lin2023video}, \cite{damonlpsg2023videollama}, \cite{2023videochat}.
		
		Surprisingly while there are many works that touch on video understanding at various levels, there have been relatively few works that have used videos in a retrieval augmented generation (RAG) \citep{Lewis2020RetrievalAugmentedGF} context; some notable related works include  EgoInstructor \cite{Xu2024RetrievalAugmentedEV} where the authors introduce a retrieval augmented multimodal captioning model that retrieves relevant exocentric videos as references to generate the captions for egocentric videos, also in \citep{10350438} they use retrieved text to generate answers for questions about an input video. Additionally in \citep{Zhang2024MagicLens} the authors improve multimodal query (image + text) to image retrieval using a large scale (query image, instruction, target image) triplet dataset. Those respective applications are great, but in this work we focus on bringing the context of a large video corpus itself into responses of a retrieval augmented generation chat bot setting. 
		
		One potential reason for relatively few works in this space is that accessing videos in a large scale can be a daunting engineering endeavor, given video information's relative large size and multimodal nature. Another potential reason for a lack of related works can be attributed to the relative difficulty of collecting video data, and further compounded by the time consuming effort of evaluating the retrieval and generation stages' output manually.
		
		In this work, we propose the use of "aligned visual caption" transcripts (see example in Figure \ref{fig:transcript}) in the context of a chat assistant. In Section \ref{sec:vc} we detail the process of preparing aligned video captions, then describe a video data set we curated for this work, and provide commentary on how these compare to using different signals from videos in conjunction with popular LLMs under the task of video summarization as a proxy for the model's capabilities in general video understanding. Then in Section \ref{sec:science} we describe an experiment that aims to automatically measure feasibility of using these transcripts in a retrieval augmented generation context. Then in Section \ref{sec:app} we describe a sample AI chat application architecture that leverages the aligned video caption representation of videos to illustrate the ease of integration. For sample demo application, LLM prompts, evaluation scripts and dataset pointers, see: \url{https://github.com/kdr/videoRAG-mrr2024}
		
		\section{Aligned Video Captions}
		\label{sec:vc}
		
\begin{figure*}
			\centering
			\includegraphics[width=\dimexpr\textwidth-2\fboxrule-2\fboxsep]{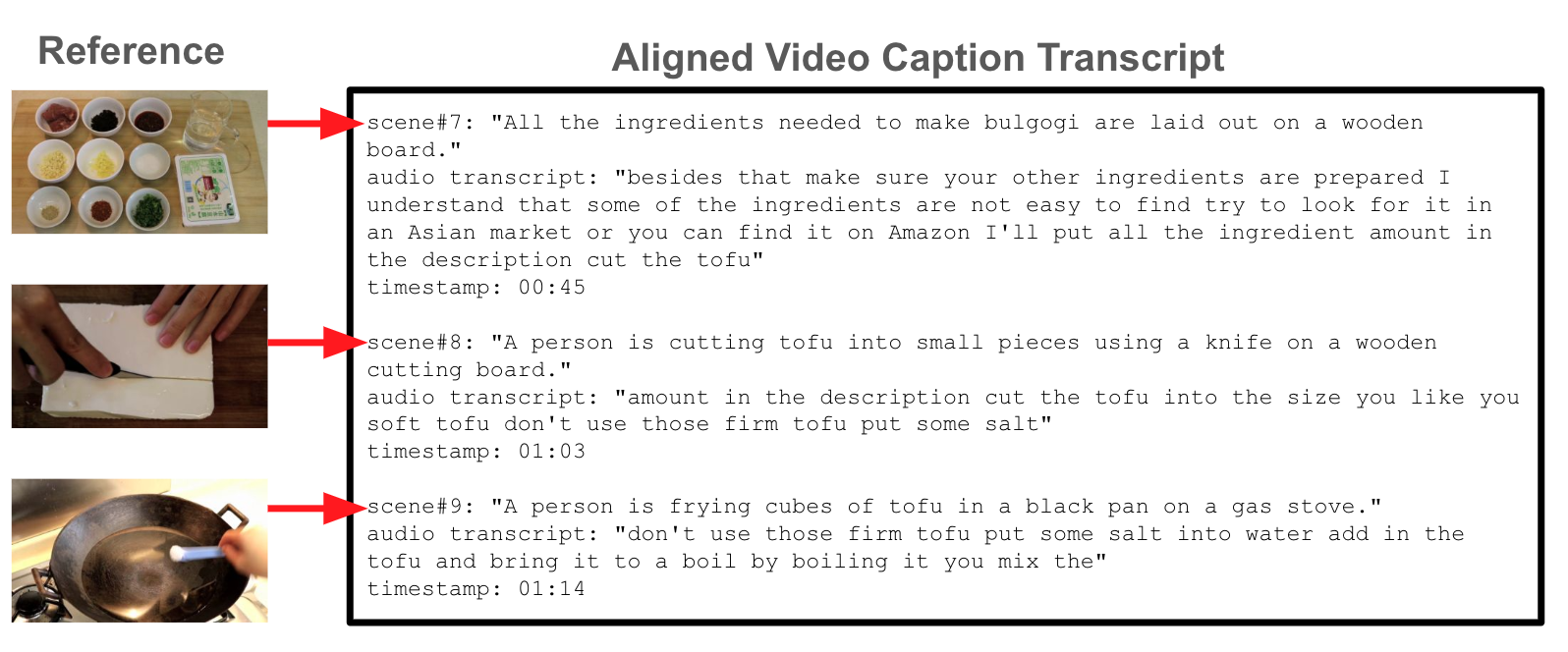}%
			\caption{Sample aligned video caption transcript with corresponding example video frames from source scenes}
			\label{fig:transcript}
		\end{figure*}

		"Aligned Video Caption Transcripts" are temporally synced scene descriptions of a video in the form of machine generated visual captions and the associated subtitles or automatic speech recognition transcripts. In this study we curated a dataset based on public youtube videos sampled from Panda-70M \cite{chen2024panda70m}, which contains individual clip segments and a general visual scene caption learned from a set of open source video captioning models. Specifically we sampled roughly 2,000 videos from each YouTube category present in Panda-70M  ~\citep{chen2024panda70m}, resulting in a dataset of 29,259 videos (1.5M video clips and corresponding visual captions) or roughly 215 days of footage. We then augment that dataset with subtitles gathered directly from YouTube's APIs and created the aligned transcripts as seen in Figure \ref{fig:transcript}).  General statistics shown in Table~\ref{tab:dataset}.
		
		In order to verify that the information an LLM can generate from an aligned video caption transcript is roughy comparable to that of a multimodal LLM, as a sanity check we checked how semantically similar video summarizations generated by various LLMs were to those generated by GPT-4 Turbo using the aligned video caption transcript. We compared these generated summaries using BERTScore \cite{bert-score}, which is an automatic summarization measure has been shown to correlate with human judgment on sentence-level and system-level evaluation. A total of 1.5K videos were summarized and evaluated, sampled uniformly from the original dataset. 
		
		In Table \ref{tab:summarization} we can see that the various configurations correlate significantly with the GPT-4 based ground truth. In particular we see that sending raw video frames and the automatic speech recognition (ASR) transcript to the GPT-4 scores a high BERTScore; so does the text only based settings using ASR, suggesting much of the information that the LLM is able to tap into resides in speech. Additionally we see that summarizations using Gemini 1.5 Pro with video based input and GPT 4 using video frames (i.e. first frame per scene) as input have similar scores as well, showing that these captions can produce similar quality output with having to send the entire set of frames to the LLM, greatly saving on context window and processing bandwidth at query time. For example, if the entire aligned video caption dataset was sampled at 1 frame per second (as is the case for popular LLMs like Gemini 1.5 pro) and assuming an image is resized to roughly fit the cost of 256 tokens for the LLM, you're looking at around 4.8 billion tokens including subtitles (roughly 69x bigger compared to using aligned visual captions). Note that Gemini 1.5 pro did not process audio signals in videos at the time this study was conducted.
		
		\begin{table}[h]
			\centering
			\begin{tabular}{l l l }
				\toprule
				\textbf{DATASET DIMENSION} & \textbf{TOTAL} & \textbf{MEDIAN} \\ \hline
				Video Count & 29,259 & - \\
				Scene Count & 1,476,462 & 31.00 \\
				Video Duration (seconds) & 18,584,396 & 478.00 \\
				\midrule		
				\multicolumn{3}{c}{\textbf{Text Character Length}} \\
				\midrule
				Title & 1,548,810 & 51.00 \\
				Description & 30,565,705 & 780.00 \\
				Title + Description & 32,114,515 & 833.00 \\
				Visual Video Captions & 96,888,187 & 2,016.00 \\
				Subtitles / ASR & 141,926,062 & 3,472.00 \\
				Aligned Captions & 276,019,918 & 6,461.00 \\
				\bottomrule
			\end{tabular}
			\caption{Statistics for Aligned Video Caption Dataset}
			\label{tab:dataset}
		\end{table}
		
		\begin{table}
			\begin{center}
				\begin{tabular}{ l l l }
					\toprule
					\multicolumn{1}{c}{\bf LLM}  &\multicolumn{1}{c}{\bf PROMPT CONTEXT} &\multicolumn{1}{c}{\bf BERT}\\
					\hline
					\multicolumn{3}{c}{\textbf{Multimodal LLMs}} \\
					\midrule
					GPT 4 V	& Video Frames + ASR Transcript	&0.889 \\
					Gemini 1.5 Pro	& Original Video	&0.862 \\
					GPT4 V & Video Frames &0.860 \\
					\midrule		
					\multicolumn{3}{c}{\textbf{GPT 4 Turbo Varying Text Input}} \\
					\midrule
					GPT 4	& ASR Transcript	&0.893 \\
					GPT 4	& Visual Captions	&0.869 \\
					GPT 4	& Title + Description	&0.858 \\
					\bottomrule
				\end{tabular}
			\end{center}
			\caption{Generated video summary comparison against GPT 4 aligned visual captions based generation}
			\label{tab:summarization}
		\end{table}
		
		\section{Video Retrieval Augmented Generation}
		\label{sec:science}

		In this section we determine if text embeddings over video derived data is feasible input for retrieval augmented generation, over the task of answering a provided general knowledge question using answers found in videos as support. In this experiment we use 1000 general knowledge questions generated via GPT 4 V as input to an embedding extractor. We also compare retrieval results using two multimodal embeddings, namely BLIP-2's ~\citep{li2023blip2} image feature extractor and CLIP ~\citep{Radford2021LearningTV} embeddings (ViT-L/14@336px). Then we retrieve the top K results as determined by a simple cosine similarity against the entire 29K video dataset. Using the top K results we use GPT-4 as an automatic judge using the following metrics:
		\begin{itemize}
			\item \textbf{HIT@K}: in the top K retrieved results, does any retrieved document contain the information required to answer the posed question. We use this in lieu of recall given the difficulty of manually collecting ground truth over every video (also answers are free form sentences and can't simply be checked for existence via basic string comparisons)
			\item \textbf{QUALITY@1}: answer correctness / quality rating between 1-10, measuring quality of answers generated by GPT-3.5 turbo. In order to control for compounding factors due to the provided context in the LLM prompt, all answers were generated using the aligned video caption transcript of the retrieved result regardless of retrieval method
		\end{itemize}
		
		To generate the questions we first sampled 500 videos from the dataset, then provided the aligned video captions as context to GPT 4 and asked the LLM to generate general knowledge questions that the video could help answer but are not specifically tied to the source video, and from the resulting question set we uniformly sampled 1000 questions.
		
		In Table \ref{tab:retrieval} we can see that the text embeddings are able to find hits at a relatively low K using the aligned transcript and ASR. We also see that the relevance of results at very low K suffers for the cross-modal embedding configuration, but can ultimately catch up if you have a tolerance for higher K, i.e. LLM can handle processing more retrieved documents in its context window, which is encouraging for future extensions into mutlimodal querying.
		
		\begin{table*}
			\centering
			\begin{tabular}{llllll}
				\toprule
				\textbf{EMBEDDING} & \textbf{DATABASE} & \textbf{HIT@1} & \textbf{HIT@5} & \textbf{HIT@10} & \textbf{ QUALITY@1} \\ \hline
				\multicolumn{6}{c}{\textbf{Multimodal Embeddings: Cross-modal Text to Vision Match}} \\
				\midrule		
				BLIP-2 & Video Frames & 0.482 & 0.801 & 0.895 & 5.199 \\
				BLIP-2 & Video Thumbnail & 0.519 & 0.833 & 0.902 & 5.598 \\
				CLIP ViT-L/14@336px & Video Frames & 0.542 & 0.858 & 0.925 & 5.785 \\
				CLIP ViT-L/14@336px & Video Thumbnail & 0.553 & 0.859 & 0.926 & 5.889 \\
				\midrule		
				\multicolumn{6}{c}{\textbf{Text Embeddings}} \\
				\midrule
				text-embedding-3-small & ASR & 0.741 & 0.936 & 0.969 & 7.424 \\
				text-embedding-3-small & Visual Captions & 0.65 & 0.878 & 0.932 & 6.605 \\
				text-embedding-3-small & Title & 0.629 & 0.905 & 0.95 & 6.503 \\
				text-embedding-3-small & Title + Description & 0.675 & 0.914 & 0.95 & 6.828 \\
				text-embedding-3-small & Aligned Transcript & 0.741 & 0.934 & 0.971 & 7.377 \\ 
				\bottomrule
			\end{tabular}
			\caption{Video retrieval results and average quality of answer generated using aligned visual action of top retrieved document} 
			\label{tab:retrieval}
		\end{table*}
		
		\begin{figure*}[h]
			\centering
			\includegraphics[width=\dimexpr\textwidth-2\fboxrule-2\fboxsep]{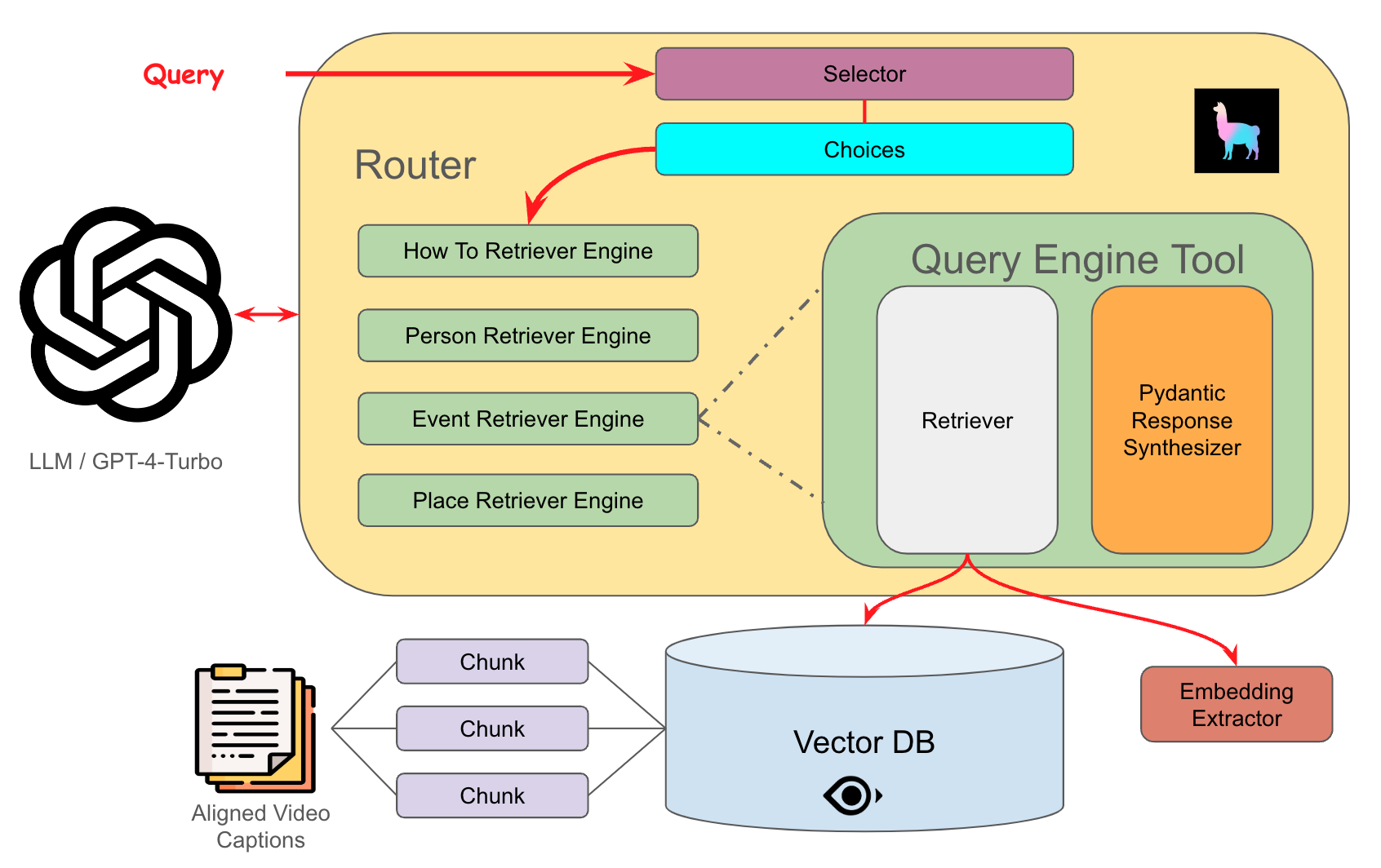}
			\caption{Example application architecture for integrating aligned video captions to enabled video enriched RAG}
			\label{fig:system}
		\end{figure*}
		
		\section{Video Enriched Chat Bot}
		\label{sec:app}
		
		In Figure \ref{fig:system} we illustrate the main components in a RAG based AI chat bot application that leverages the aligned video captions to return relevant answers and corresponding video clip source. The processing is as follows:
		
		\begin{enumerate}
			\item Based on the user query and a choice of tool descriptions, one retriever tool is selected; we created different tools that point to specific subsets of the video catalog
			\item The selected query engine tool vectorizes the query and searches the vector database to retrieve (chunked) aligned video caption text blobs.
			\item The query engine tool interprets the results and summarizes into a specific pydantic format customized for that answer type; for example a "how to" response should respond with a bulleted list of steps like in Figure \ref{fig:teaser}, whereas a "place" response would describe a location and why it is notable. Timestamps in retrieved docs help give the application pointers to specific parts of video to enhance user interaction
		\end{enumerate}
		
		\section{Closing Remarks}
		\label{sec:closing}
		
		In this study we show that aligned visual captions provide a compelling and adaptable representation of video information that can easily plug into basic LLM application architectures. We curate a large scale dataset, demonstrate how to leverage the data representation to generate questions, and offer an automated procedure for measuring video RAG based question answering results.
		
		This work gives us a glimpse into the potential of using aligned video captions representation, and is ripe for future exploration. For example, a key practical consideration in deploying this solution in the real world is the availability of video captioning models suitable for the intended use case. Another thing to consider is how one identifies meaningful video clip segments to be summarized by the captioning models in the first place. In future works it would be interesting to study how generic video captioning and clip segmentation methods fare on different video domains (e.g. general knowledge vs. surveillance, etc.) and contrast those with strategies that adapt the various components in the processing pipeline for a target domain. Moreover, the audio signal incorporated in this study focuses on the spoken word, and for other domains it may be interesting to incorporate other aspects of the audio as well (e.g. descriptions of music or shifts in loudness could give hints to the overall tone or emotion of the scene).

		\bibliographystyle{ACM-Reference-Format}
		\bibliography{ref}


\begin{thebibliography}{12}


\ifx \showCODEN    \undefined \def \showCODEN     #1{\unskip}     \fi
\ifx \showDOI      \undefined \def \showDOI       #1{#1}\fi
\ifx \showISBNx    \undefined \def \showISBNx     #1{\unskip}     \fi
\ifx \showISBNxiii \undefined \def \showISBNxiii  #1{\unskip}     \fi
\ifx \showISSN     \undefined \def \showISSN      #1{\unskip}     \fi
\ifx \showLCCN     \undefined \def \showLCCN      #1{\unskip}     \fi
\ifx \shownote     \undefined \def \shownote      #1{#1}          \fi
\ifx \showarticletitle \undefined \def \showarticletitle #1{#1}   \fi
\ifx \showURL      \undefined \def \showURL       {\relax}        \fi
\providecommand\bibfield[2]{#2}
\providecommand\bibinfo[2]{#2}
\providecommand\natexlab[1]{#1}
\providecommand\showeprint[2][]{arXiv:#2}

\bibitem[Chen et~al\mbox{.}(2024)]%
        {chen2024panda70m}
\bibfield{author}{\bibinfo{person}{Tsai-Shien Chen},
  \bibinfo{person}{Aliaksandr Siarohin}, \bibinfo{person}{Willi Menapace},
  \bibinfo{person}{Ekaterina Deyneka}, \bibinfo{person}{Hsiang-wei Chao},
  \bibinfo{person}{Byung~Eun Jeon}, \bibinfo{person}{Yuwei Fang},
  \bibinfo{person}{Hsin-Ying Lee}, \bibinfo{person}{Jian Ren},
  \bibinfo{person}{Ming-Hsuan Yang}, {and} \bibinfo{person}{Sergey Tulyakov}.}
  \bibinfo{year}{2024}\natexlab{}.
\newblock \showarticletitle{Panda-70M: Captioning 70M Videos with Multiple
  Cross-Modality Teachers}.
\newblock \bibinfo{journal}{\emph{arXiv preprint arXiv:2402.19479}}
  (\bibinfo{year}{2024}).
\newblock


\bibitem[Lewis et~al\mbox{.}(2020)]%
        {Lewis2020RetrievalAugmentedGF}
\bibfield{author}{\bibinfo{person}{Patrick Lewis}, \bibinfo{person}{Ethan
  Perez}, \bibinfo{person}{Aleksandara Piktus}, \bibinfo{person}{Fabio
  Petroni}, \bibinfo{person}{Vladimir Karpukhin}, \bibinfo{person}{Naman
  Goyal}, \bibinfo{person}{Heinrich Kuttler}, \bibinfo{person}{Mike Lewis},
  \bibinfo{person}{Wen tau Yih}, \bibinfo{person}{Tim Rockt{\"a}schel},
  \bibinfo{person}{Sebastian Riedel}, {and} \bibinfo{person}{Douwe Kiela}.}
  \bibinfo{year}{2020}\natexlab{}.
\newblock \showarticletitle{Retrieval-Augmented Generation for
  Knowledge-Intensive NLP Tasks}.
\newblock \bibinfo{journal}{\emph{ArXiv}}  \bibinfo{volume}{abs/2005.11401}
  (\bibinfo{year}{2020}).
\newblock
\urldef\tempurl%
\url{https://api.semanticscholar.org/CorpusID:218869575}
\showURL{%
\tempurl}


\bibitem[Li et~al\mbox{.}(2023b)]%
        {li2023blip2}
\bibfield{author}{\bibinfo{person}{Junnan Li}, \bibinfo{person}{Dongxu Li},
  \bibinfo{person}{Silvio Savarese}, {and} \bibinfo{person}{Steven Hoi}.}
  \bibinfo{year}{2023}\natexlab{b}.
\newblock \showarticletitle{{BLIP-2:} Bootstrapping Language-Image Pre-training
  with Frozen Image Encoders and Large Language Models}. In
  \bibinfo{booktitle}{\emph{ICML}}.
\newblock


\bibitem[Li et~al\mbox{.}(2023a)]%
        {2023videochat}
\bibfield{author}{\bibinfo{person}{Kunchang Li}, \bibinfo{person}{Yinan He},
  \bibinfo{person}{Yi Wang}, \bibinfo{person}{Yizhuo Li},
  \bibinfo{person}{Wenhai Wang}, \bibinfo{person}{Ping Luo},
  \bibinfo{person}{Yali Wang}, \bibinfo{person}{Limin Wang}, {and}
  \bibinfo{person}{Yu Qiao}.} \bibinfo{year}{2023}\natexlab{a}.
\newblock \showarticletitle{VideoChat: Chat-Centric Video Understanding}.
\newblock \bibinfo{journal}{\emph{arXiv preprint arXiv:2305.06355}}
  (\bibinfo{year}{2023}).
\newblock


\bibitem[Lin et~al\mbox{.}(2023)]%
        {lin2023video}
\bibfield{author}{\bibinfo{person}{Bin Lin}, \bibinfo{person}{Bin Zhu},
  \bibinfo{person}{Yang Ye}, \bibinfo{person}{Munan Ning},
  \bibinfo{person}{Peng Jin}, {and} \bibinfo{person}{Li Yuan}.}
  \bibinfo{year}{2023}\natexlab{}.
\newblock \showarticletitle{Video-LLaVA: Learning United Visual Representation
  by Alignment Before Projection}.
\newblock \bibinfo{journal}{\emph{arXiv preprint arXiv:2311.10122}}
  (\bibinfo{year}{2023}).
\newblock


\bibitem[Maaz et~al\mbox{.}(2023)]%
        {Maaz2023VideoChatGPT}
\bibfield{author}{\bibinfo{person}{Muhammad Maaz}, \bibinfo{person}{Hanoona
  Rasheed}, \bibinfo{person}{Salman Khan}, {and} \bibinfo{person}{Fahad~Shahbaz
  Khan}.} \bibinfo{year}{2023}\natexlab{}.
\newblock \showarticletitle{Video-ChatGPT: Towards Detailed Video Understanding
  via Large Vision and Language Models}.
\newblock \bibinfo{journal}{\emph{arXiv:2306.05424}} (\bibinfo{year}{2023}).
\newblock


\bibitem[Pan et~al\mbox{.}(2023)]%
        {10350438}
\bibfield{author}{\bibinfo{person}{J. Pan}, \bibinfo{person}{Z. Lin},
  \bibinfo{person}{Y. Ge}, \bibinfo{person}{X. Zhu}, \bibinfo{person}{R.
  Zhang}, \bibinfo{person}{Y. Wang}, \bibinfo{person}{Y. Qiao}, {and}
  \bibinfo{person}{H. Li}.} \bibinfo{year}{2023}\natexlab{}.
\newblock \showarticletitle{Retrieving-to-Answer: Zero-Shot Video Question
  Answering with Frozen Large Language Models}. In
  \bibinfo{booktitle}{\emph{2023 IEEE/CVF International Conference on Computer
  Vision Workshops (ICCVW)}}. \bibinfo{publisher}{IEEE Computer Society},
  \bibinfo{address}{Los Alamitos, CA, USA}, \bibinfo{pages}{272--283}.
\newblock
\urldef\tempurl%
\url{https://doi.org/10.1109/ICCVW60793.2023.00035}
\showDOI{\tempurl}


\bibitem[Radford et~al\mbox{.}(2021)]%
        {Radford2021LearningTV}
\bibfield{author}{\bibinfo{person}{Alec Radford}, \bibinfo{person}{Jong~Wook
  Kim}, \bibinfo{person}{Chris Hallacy}, \bibinfo{person}{Aditya Ramesh},
  \bibinfo{person}{Gabriel Goh}, \bibinfo{person}{Sandhini Agarwal},
  \bibinfo{person}{Girish Sastry}, \bibinfo{person}{Amanda Askell},
  \bibinfo{person}{Pamela Mishkin}, \bibinfo{person}{Jack Clark},
  \bibinfo{person}{Gretchen Krueger}, {and} \bibinfo{person}{Ilya Sutskever}.}
  \bibinfo{year}{2021}\natexlab{}.
\newblock \showarticletitle{Learning Transferable Visual Models From Natural
  Language Supervision}. In \bibinfo{booktitle}{\emph{International Conference
  on Machine Learning}}.
\newblock
\urldef\tempurl%
\url{https://api.semanticscholar.org/CorpusID:231591445}
\showURL{%
\tempurl}


\bibitem[Xu et~al\mbox{.}(2024)]%
        {Xu2024RetrievalAugmentedEV}
\bibfield{author}{\bibinfo{person}{Jilan Xu}, \bibinfo{person}{Yifei Huang},
  \bibinfo{person}{Junlin Hou}, \bibinfo{person}{Guo Chen},
  \bibinfo{person}{Yue Zhang}, \bibinfo{person}{Rui Feng}, {and}
  \bibinfo{person}{Weidi Xie}.} \bibinfo{year}{2024}\natexlab{}.
\newblock \showarticletitle{Retrieval-Augmented Egocentric Video Captioning}.
\newblock \bibinfo{journal}{\emph{ArXiv}}  \bibinfo{volume}{abs/2401.00789}
  (\bibinfo{year}{2024}).
\newblock
\urldef\tempurl%
\url{https://api.semanticscholar.org/CorpusID:266693245}
\showURL{%
\tempurl}


\bibitem[Zhang et~al\mbox{.}(2023)]%
        {damonlpsg2023videollama}
\bibfield{author}{\bibinfo{person}{Hang Zhang}, \bibinfo{person}{Xin Li}, {and}
  \bibinfo{person}{Lidong Bing}.} \bibinfo{year}{2023}\natexlab{}.
\newblock \showarticletitle{Video-LLaMA: An Instruction-tuned Audio-Visual
  Language Model for Video Understanding}.
\newblock \bibinfo{journal}{\emph{arXiv preprint arXiv:2306.02858}}
  (\bibinfo{year}{2023}).
\newblock
\urldef\tempurl%
\url{https://arxiv.org/abs/2306.02858}
\showURL{%
\tempurl}


\bibitem[Zhang et~al\mbox{.}(2024)]%
        {Zhang2024MagicLens}
\bibfield{author}{\bibinfo{person}{Kai Zhang}, \bibinfo{person}{Yi Luan},
  \bibinfo{person}{Hexiang Hu}, \bibinfo{person}{Kenton Lee},
  \bibinfo{person}{Siyuan Qiao}, \bibinfo{person}{Wenhu Chen},
  \bibinfo{person}{Yu Su}, {and} \bibinfo{person}{Ming-Wei Chang}.}
  \bibinfo{year}{2024}\natexlab{}.
\newblock \showarticletitle{MagicLens: Self-Supervised Image Retrieval with
  Open-Ended Instructions}.
\newblock \bibinfo{journal}{\emph{arXiv preprint arXiv:2403.19651}}
  (\bibinfo{year}{2024}).
\newblock


\bibitem[Zhang* et~al\mbox{.}(2020)]%
        {bert-score}
\bibfield{author}{\bibinfo{person}{Tianyi Zhang*}, \bibinfo{person}{Varsha
  Kishore*}, \bibinfo{person}{Felix Wu*}, \bibinfo{person}{Kilian~Q.
  Weinberger}, {and} \bibinfo{person}{Yoav Artzi}.}
  \bibinfo{year}{2020}\natexlab{}.
\newblock \showarticletitle{BERTScore: Evaluating Text Generation with BERT}.
  In \bibinfo{booktitle}{\emph{International Conference on Learning
  Representations}}.
\newblock
\urldef\tempurl%
\url{https://openreview.net/forum?id=SkeHuCVFDr}
\showURL{%
\tempurl}


\end{thebibliography}
		
	\end{document}